\documentclass[10pt, a4paper]{article}
\usepackage{lrec2022} 
\usepackage{multibib}
\newcites{languageresource}{Language Resources}
\usepackage{graphicx}
\usepackage{tabularx}
\usepackage{soul}
\usepackage{booktabs}
\usepackage{titlesec}
\titleformat{\section}{\normalfont\large\bf\center}{\thesection.}{1em}{}
\titleformat{\subsection}{\normalfont\SmallTitleFont\bf\raggedright}{\thesubsection.}{1em}{}
\titleformat{\subsubsection}{\normalfont\normalsize\bf\raggedright}{\thesubsubsection.}{1em}{}
\renewcommand\thesection{\arabic{section}}
\renewcommand\thesubsection{\thesection.\arabic{subsection}}
\renewcommand\thesubsubsection{\thesubsection.\arabic{subsubsection}}
\usepackage{epstopdf}
\usepackage[utf8]{inputenc}
\usepackage{xstring}
\usepackage{color}
\usepackage[hidelinks]{hyperref}

\title{The Copenhagen Corpus of Eye Tracking Recordings from Natural Reading of Danish Texts}

\name{Nora Hollenstein\textsuperscript{1}, Maria Barrett\textsuperscript{2}, Marina Björnsdóttir\textsuperscript{1,2}}

\address{\textsuperscript{1} Center for Language Technology, University of Copenhagen \\
\textsuperscript{2} Computer Science Department, IT University of Copenhagen \\
nora.hollenstein@hum.ku.dk, mbarrett@itu.dk, marina.bjorns@gmail.com
\\}

\abstract{Eye movement recordings from reading are one of the richest signals of human language processing. Corpora of eye movements during reading of contextualized running text is a way of making such records available for natural language processing purposes. Such corpora already exist in some languages. We present CopCo, the Copenhagen Corpus of eye tracking recordings from natural reading of Danish texts. It is the first eye tracking corpus of its kind for the Danish language. CopCo includes 1,832 sentences with 34,897 tokens of Danish text extracted from a collection of speech manuscripts. This first release of the corpus contains eye tracking data from 22 participants. It will be extended continuously with more participants and texts from other genres. We assess the data quality of the recorded eye movements and find that the extracted features are in line with related research. The dataset available here: \url{https://osf.io/ud8s5/}.
\\
 \newline\newline \Keywords{eye tracking, reading, Danish, corpus, psycholinguistics, cognitively-inspired NLP} }

\begin{document}

\maketitleabstract

\section{Introduction}
Records of eye movements during reading have been studied since decades in controlled psycholinguistic studies. The advantages of eye movement data for studying cognitive language processing are well established. For example, there is a thoroughly examined link between the fixation duration on a word and the cognitive effort required to process this word. During skilled reading, the eyes fixate sequentially through the text but approximately 10--15\% of the fixations occur in a previously read part of the text to further process it. Most often, the reader is not aware of this. Therefore, eye movements allow us to study both early and later stages of cognitive text processing \cite{rayner1989eye}.\footnote{For a review, see \newcite{clifton2007eye}.}

In controlled studies, participants typically read constructed individual sentences. Often, pairs of sentences with minimal differences are used to study one particular linguistic phenomenon \cite{brennan2016naturalistic}. However, as highlighted by \newcite{demberg2019cognitive}, there is a need for evaluating psycholinguistic theories on natural reading corpora rather than on highly infrequent sentence constructions specifically designed for the purpose of one experiment. Eye tracking and brain activity studies during unpaced reading of naturally occurring, contextualized text have recently gained more attention in the research community thanks to advances in recording technologies \cite{hamilton2018revolution,sato2018successful}. While natural eye tracking corpora cannot replace controlled psycholinguistic studies, it is a complementary method of studying some of the same phenomena with higher ecological validity. A naturalistic reading setup allows us to analyze real-time language processing from real-world text \cite{hasson2015naturalistic}.

\begin{table*}[ht]
\centering
\begin{tabular}{lrrrrrcc}
\toprule
\textbf{Speech ID}    & \textbf{Sents.} & \textbf{Tokens} & \textbf{Types} & \textbf{Sent. length}      & \textbf{Token length}  & \textbf{Freq.} & \textbf{LIX}   \\\midrule
1125  & 132               & 1917                & 611                & 14.52 (8.75, 1-45)   & 3.84 (2.87, 1-22) & 0.74            & 30.16 \\
1165  & 71                & 1319                & 454                & 18.58 (11.5, 1-59)   & 3.71 (2.39, 1-17) & 0.78            & 30.12 \\
1317  & 107               & 1830                & 683                & 17.1 (7.53, 5-44)    & 4.73 (3.86, 1-34) & 0.74      & 43.19 \\
1318  & 114    & 2143    & 711         & 18.8 (9.73, 5-51)    & 4.38 (3.39, 1-25) & 0.75            & 41.49 \\
1323  & 100               & 2044                & 720                & 20.44 (10.59, 4-50)  & 4.62 (3.7, 1-25)  & 0.74            & 44.52 \\
7797  & 119               & 1639                & 645                & 13.77 (8.63, 1-37)   & 3.94 (2.96, 1-24) & 0.70            & 31.00 \\
7856  & 59                & 2139                & 634                & 36.25, (26.47, 2-139) & 3.75 (2.55, 1-21) & 0.77            & 45.58 \\
7905  & 106               & 2648                & 1056               & 24.98 (14.74, 3-67)  & 4.5, (3.3, 1-24)   & 0.73            & 48.58 \\
7946  & 134               & 1700                & 591                & 12.69 (9.72, 1-80)   & 3.82 (2.74, 1-20) & 0.71            & 26.43 \\
10365 & 126               & 1782                & 615                & 14.14, (8.05, 2-43)   & 3.69 (2.64, 1-24) & 0.71            & 26.77 \\
10440 & 89                & 1150                & 409                & 12.92, (7.64, 2-38)   & 3.98 (3.23, 1-22) & 0.73            & 29.08 \\
11171 & 51                & 1170                & 477                & 22.94 (13.78, 1-56)  & 3.97 (2.86, 1-20) & 0.73            & 39.09 \\
12063 & 57                & 1172                & 498                & 20.56 (14.26, 2-57)  & 3.89 (2.81, 1-23) & 0.68            & 34.79 \\
17526 & 109               & 2846                & 784                & 26.11 (15.91, 1-71)  & 3.8, (2.85, 1-24)  & 0.74            & 39.79 \\
18473 & 37                & 978                 & 391                & 26.43 (13.98, 4-58)  & 4.51 (3.56, 1-20) & 0.75            & 50.14 \\
18561 & 82                & 1260                & 410                & 15.37 (7.91, 1-37)   & 3.87, (2.93, 1-18) & 0.74            & 31.70 \\
18670 & 81                & 1282                & 480                & 15.83 (9.38, 1-54)   & 4.05 (3.09, 1-21) & 0.72            & 34.81 \\
22811 & 54                & 1357                & 524                & 25.13 (19.4, 1-97)   & 4.16 (3.41, 1-29) & 0.73            & 41.49 \\
26670 & 102               & 2215                & 641                & 21.72 (11.7, 1-52)   & 3.94 (2.73, 1-22) & 0.79            & 37.79 \\
26682 & 119               & 2306                & 737                & 19.38 (10.66, 1-52)  & 4.1, (2.93, 1-21)  & 0.78            & 37.85\\\midrule
total & 1832 & 34897 & 5872 & 19.05 (13.07, 1-139) & 4.07 (3.08, 1.34) & 0.74 & 37.22 \\\bottomrule
\end{tabular}
\caption{Dataset statistics. The speech ID is the original ID from the source corpus; sentence length is the mean number of tokens per sentence (with standard deviation and range in brackets); token length is the mean number of characters per token (with standard deviation and range in brackets); frequency is the proportion of words included in the 10,000 most common Danish words (Source: \url{https://korpus.dsl.dk/resources/details/freq-lemmas.html});\hspace{0.2pt} LIX is the readability score as described in Section \ref{sec:materials}.}
\label{tab:speech-stats}
\end{table*}

An obvious advantage of larger corpora of naturally occurring text is that the same corpus can be reused for multiple purposes. Researchers can either isolate the relevant linguistic phenomenon or model the entire reading process. Instead of comparing binarized groups of words, e.g., low-frequency and high-frequency words in a controlled study, eye tracking corpora allow us to model the entire spectrum as done by \newcite{kennedy2013frequency}. Similarly, controlled studies have shown that the syntax of a sentence is processed at the end of the sentence when comparing minimal pairs of sentences with different syntactical complexity \cite{traxler1997influence,warren2009investigating}, the so-called wrap-up effect, but eye tracking corpora can answer how punctuation influences reading in contextualized naturally occurring sentences that contain the full spectrum of simple and more complex syntactic constructions \cite{pynte2007influence}.

Another benefit of the larger natural eye movement corpora is their potential for NLP \cite{hollenstein2020towards}. Some of these corpora are large enough to train cognitively inspired models. The growing list of gaze-augmented NLP models includes tasks such as image captioning \cite{takmaz2020generating}, named entity recognition \cite{hollenstein2019entity}, sentiment analysis \cite{mishra2016leveraging}, or part-of-speech tagging \cite{barrett2016weakly}.\footnote{See \newcite{mathias2020survey} and \newcite{barrett2020sequence} for more extensive reviews.} More recently, eye tracking data from reading has also been leveraged to evaluate and interpret computational language models \cite{sood2020interpreting,abdou2019higher,hollenstein-beinborn-2021-relative}.

There are eye tracking corpora from natural reading available in other languages, but as of yet, no such resource is available for Danish. We present CopCo, the Copenhagen Eye Tracking Corpus. It is the first Danish eye tracking corpus with contextualized, running text and self-paced reading. CopCo is a growing dataset, and the first release includes recordings from 22 participants over more than 30,000 tokens.
CopCo is freely available here: \url{https://osf.io/ud8s5/}.

\section{Related Work}
Some of the existing corpora emerged from psycholinguistic experiments, while others were tailored for their use in natural language processing applications.
In English, there are a number of eye tracking corpora of skilled adult readers performing self-paced reading of contextualized text, e.g., \newcite{kennedy2003dundee}, \newcite{cop2017presenting}, \newcite{luke2018provo}, \newcite{mishra2016predicting}. Some encompass more than 50,000 tokens, while others are smaller and focus on individual sentence processing \cite{frank2013reading,hollenstein2018zuco,hollenstein2020zuco}. Several participants read the same text; in the Provo Corpus as many as 470 \cite{luke2018provo}, but in the remaining cited corpora, the number of subjects reading the same text is between 10 and 20. 

Natural eye tracking corpora also exist in other languages.
Some of these are again smaller and focus on individual sentence processing: \newcite{husain2015integration} for Hindi, \newcite{safavi2016dependency} for Persian, \newcite{laurinavichyute2019russian} for Russian, and \newcite{pan2021beijing} for Chinese. The most recent release is the Multilingual Eye Movement Corpus (MECO; \cite{kuperman2022text,siegelman2022expanding}), a resource that includes parallel data from 580 readers in 13 different languages, reading in their native language as well as in English, following the same experiment protocol. However, this corpus does not include Danish.
Others study specific linguistic aspects, for example, \newcite{cop2017presenting} study bilingual reading processing of Dutch and English on a full novel, \newcite{jaeger2021potsdam} analyze reading patterns between experts and non-experts. 

Finally, some of the existing resources are explicitly targeted for their use in NLP applications. For instance, \newcite{yi2020dataset} compile a Chinese dataset of gaze behavior from text summarization and \newcite{sood2021vqa} provide an English dataset of visual question answering.

CopCo is a new resource in this landscape of eye tracking corpora and provides data to analyze psycholinguistic research questions as well as NLP applications. All materials are freely available so that annotations or labels for specific NLP tasks can be added in future work.

\begin{table*}[t]
\centering
\begin{tabular}{ccccccc}
\toprule
\textbf{ID} & \textbf{Age} & \textbf{Sex} & \textbf{Comp. score} & \textbf{\# Speeches} & \textbf{\# Questions} & \textbf{Reading time} \\\midrule
P01 & 29 & F & 0.92 & 4 & 13 & 25.95 \\
P02 & 62 & F & 0.83 & 6 & 18 & 14.28 \\
P03 & 23 & F & 0.95 & 6 & 20 & 18.66 \\
P04 & 26 & F & 0.88 & 2 & 8 & 20.37 \\
P05 & 44 & F & 0.85 & 7 & 26 & 13.88 \\
P06 & 47 & M & 0.78 & 6 & 18 & 14.68 \\
P07 & 26 & F & 0.81 & 4 & 16 & 22.27 \\
P08 & 32 & M & 0.8 & 2 & 5 & 14.72 \\
P09 & 39 & F & 0.0 & 1 & 1 & 21.25 \\
P10 & 25 & F & 0.81 & 4 & 16 & 22.41 \\
P11 & 32 & F & 0.78 & 6 & 23 & 13.62 \\
P12 & 29 & M & 0.86 & 4 & 14 & 18.27 \\
P13 & 32 & M & 1.0 & 3 & 8 & 15.36 \\
P14 & 59 & F & 0.85 & 6 & 20 & 13.19 \\
P15	& 25	& F	& 0.87 &	4 &	15	& 19.13 \\
P16	& 22 & 	M	& 0.79	& 4	& 14	& 14.24\\
P17	& 37	& F	& 1.0	& 2	& 7	& 17.75 \\
P18  &26& F & 0.82  &     4   &   11  & 18.50   \\
P19  &21&F & 0.62  &   4  & 13  & 16.19    \\  
P20  &24&F &  0.87  &  6  & 23  &     10.03   \\
P21 &26 &F & 1.0 & 4 & 16 & 14.41 \\
P22 & 23 &F & 0.80 & 4 & 15 & 18.67 \\\midrule
mean & 32 & - & 0.82 & 4.13 & 14.3 & 17.27\\\bottomrule
\end{tabular}
\caption{Participant statistics. Comp. score: proportion of correctly answered reading comprehension questions; number of speeches and number of questions read/answered by this participant; absolute reading time: seconds spent on each screen.}
\label{tab:participants}
\end{table*}

\section{Experiment Design}

\subsection{Reading Materials}\label{sec:materials}
All reading materials are Danish speech manuscripts from \url{https://dansketaler.dk/}. We selected speeches limited to the following categories: event or conference speeches, speeches given by a bonfire at a solstice event, and high school graduation speeches. None of the conferences are scientific conferences for a highly specialised audience. The speeches from the selected categories are expected to be engaging for a general audience. Although speech manuscripts are not a frequently studied genre, the Danske Taler archive provides a great resource of longer, yet self-contained Danish texts of current language use without copyright. The content is largely comparable to essays; a piece of writing on a particular topic, often from a personal point of view. We sampled speeches from the years 2010--2019. This selection returned 46 speeches, and we manually subsampled them to remove speeches that are not interesting to a general audience, as well as to get a broad distribution of speaker demographics in the final sample. We balanced the gender distribution of the speakers such that the corpus contains ten male and ten female speakers.

Furthermore, our objective was to obtain as broad an age range and range of the geographical location of the speech event as possible. Some of the speeches were already proofread by Danske Taler. Nevertheless, a native Danish speaker proofread all speeches chosen for this data collection.\footnote{Not all speech manuscripts used canonical punctuation but rather marked pauses for the speaker - sometimes with hyphens. We tried to maintain the texts in their original form as far as possible. We edited clearly incorrect punctuation but did allow extensive use of hyphen instead of comma and full stop. We acknowledge that replacing full stops with hyphens inaccurately inflates the readability score.} 

In total, CopCo contains 34,897 tokens in 1,832 sentences in a selection of 20 speeches. Table \ref{tab:speech-stats} presents the statistics of the data set for each speech, including the number of words, sentences, the average word length, sentence length, and the LIX score \cite{bjornsson1968lasbarhet} as calculated by the \texttt{readcalc} Python library.\footnote{\url{https://pypi.org/project/ReadabilityCalculator/}} The LIX score is a simple readability metric considering the length of words and the length of sentences. A score of 25--34 is considered an easy text for skilled adult readers, and $>$55 is considered difficult.  

\subsection{Comprehension Questions}

\begin{figure*}[t]
\centering
\includegraphics[width=0.68\textwidth]{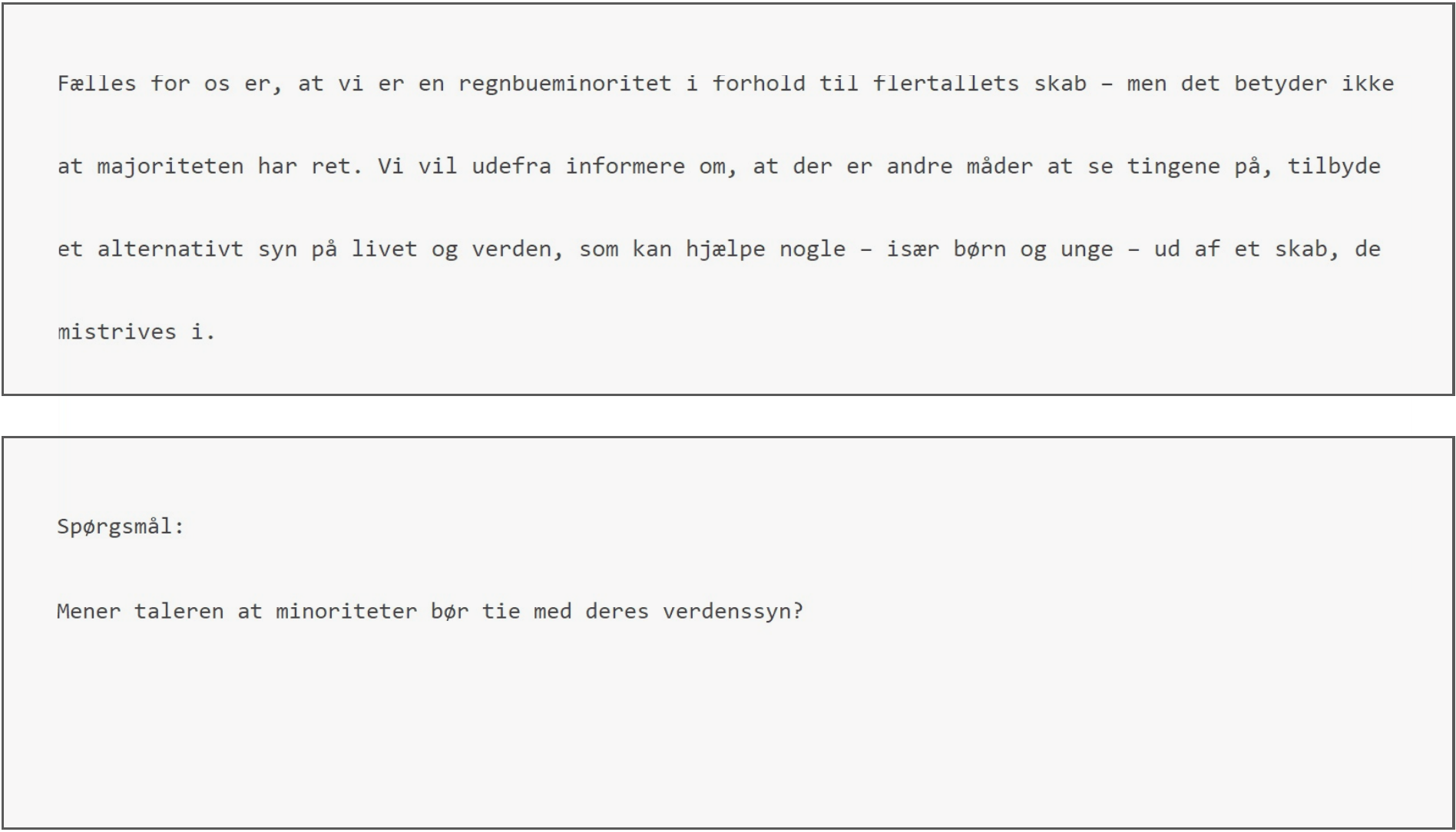}
\caption{Example screen of a text paragraph and the following question on the next screen.}\label{fig:procedure}
\end{figure*}

After approximately 20\% of all paragraphs longer than 100 characters, the subjects are presented with a comprehension question related to the previously read paragraph to prevent mindless reading. The subset of paragraphs was selected randomly. 
Each question must be answered with \textit{yes} or \textit{no}. There is no option to skip a question. The questions either ask about the factual details described in the text or a verification of a more general and shallow synthesis of the text. In total, there are 68 questions -- 30 where the correct answer is \textit{no}, 38 where the correct answer is \textit{yes}.

\subsection{Participants}
 
The participants are adult, native speakers of Danish. All have normal vision or corrected-to-normal vision (glasses or contact lenses) and no known reading impairments.
At the time of publication, we collected eye movement data from 22 participants (23\% male, between 21 and 62 years old, with the highest completed education levels ranging between high school and PhD). Participant recruitment is still in progress and data of further participants will be released as soon as available. Participation was rewarded with a symbolic gift. All participants gave written consent to their participation and the reuse of the data for research purposes prior to the start of the experiment. This study was approved by the Research Ethics Committee at the Faculty of Humanities of the University of Copenhagen. 

Table \ref{tab:participants} shows the details of the participant population. For each participant, we report age and sex, the number of speeches they read, the number of comprehension questions answered, the average reading time per screen, as well as the proportion of correctly answered comprehension questions. In total, the participants have read 95 texts. Each text has been read by 3 to 8 readers. Due to bad calibration and technical problems, the data of participant P14 are not used for any further analysis.

\subsection{Recording Procedure}

At the beginning of the experiment, participants were instructed to move as little as possible and read as naturally as possible, as they would read for comprehension outside the laboratory. Participants rested their heads on an adjustable chin rest to limit head and body movements, and they used a control pad to move to the next screen and answer the comprehension questions in their own speed. Reverting to a previous screen was not possible. The reading was self-paced, which means that participants pressed a key after finishing reading each screen to move onto the next. There was no time restriction; neither to the reading of a single screen nor to the session duration. Instructions for the task were presented orally as well as on the computer screen before the experiment start. All participants were first presented with a short speech as a practice round.

The experiment was split into blocks of two speeches. The order of the blocks and the order of the speeches within a block were randomized. The experiment design allows a flexible length of the recording sessions. Each session entails at least one block (i.e., two speeches) and if the participant is not too tired, subsequent blocks can be added. On average, participants read 4.13 speeches per session. This setup allows us to extend the corpus with additional texts in the future while maintaining consistency in the experiment procedure.

\subsection{Stimulus Presentation}

The text passages presented on each screen resembled the author’s original division of the story into paragraphs as much as possible. Comprehension questions were presented on separate screens and clearly marked with the title ``Spørgsmål" (translation: ``Question").

The text was presented in a black, monospaced font (font type: Consolas; font size: 16) on a light-gray background (RGB: 248,248,248) as shown in Figure \ref{fig:procedure}. The texts spanned multiple lines (max. 10) with triple line spacing. The text was presented with a 140 pixels margin at the top and bottom, and 200 pixels on the left and right.

\begin{figure*}[t]
\centering
\includegraphics[width=0.9\textwidth]{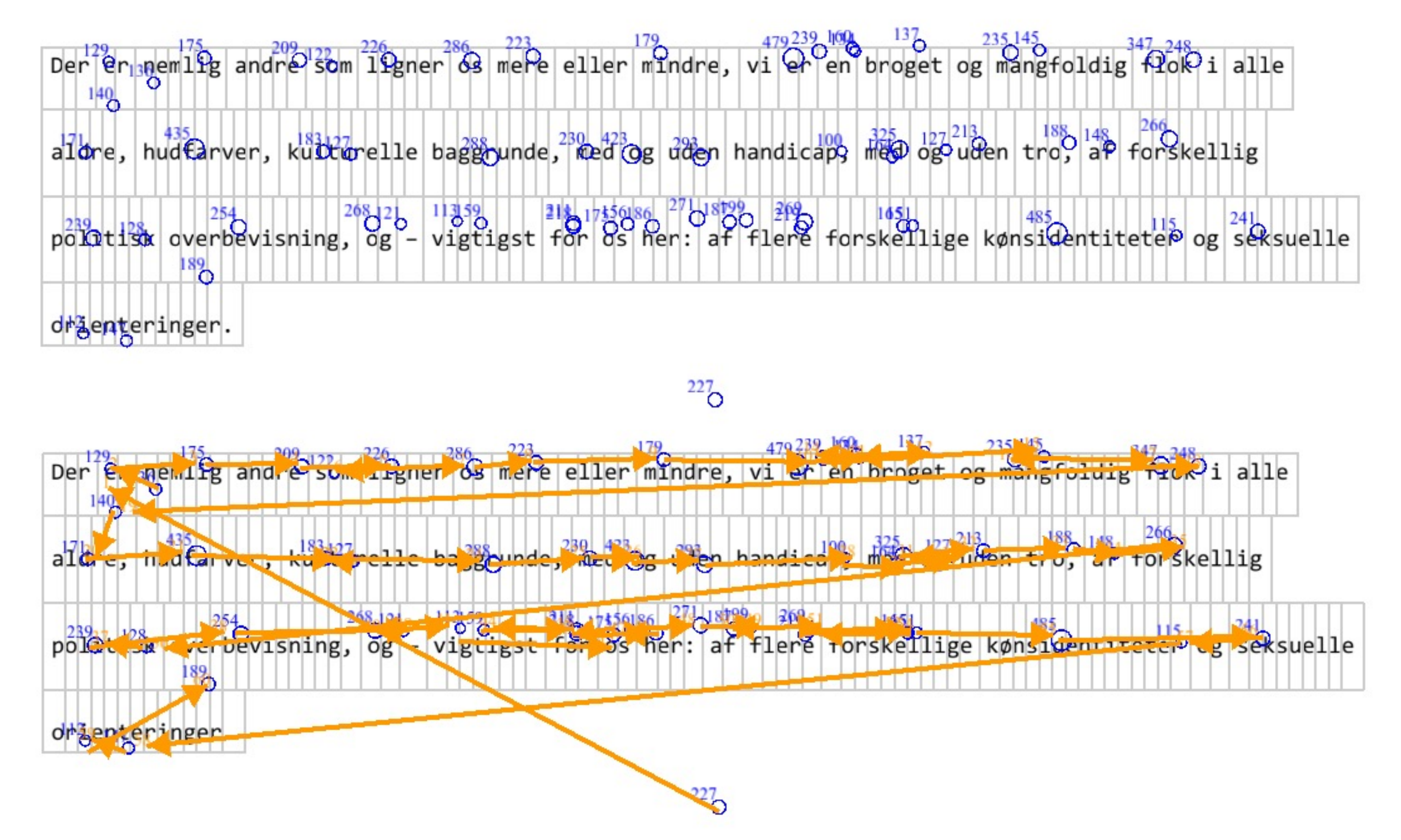}
\caption{(Top) Fixations detected from a single reader; the numbers in blue show the fixation duration in milliseconds. (Bottom) The corresponding saccades between the fixations, the arrow heads mark the direction of the eye movement.}\label{fig:fix-saccs}
\end{figure*}

\section{Data Acquisition}

Eye movement data was collected with an infrared video-based EyeLink 1000 Plus eye tracker (SR Research). The experiment was designed with the SR Experiment Builder software. Data is recorded with a sampling rate of 1000 Hz.

Participants were seated at a distance of approximately 85 cm from a 27-inch monitor (display dimensions 590 x 335 mm, resolution 1920 x 1080 pixels). We recorded monocular eye tracking data of the right eye. In a few cases of calibration difficulties, the left eye was tracked.

A 9-point calibration was performed at the beginning of the experiment. The calibration was validated after each block. Re-calibration was conducted if the quality was not good (worst point error $<1.5^{\circ}$, average error $<1.0^{\circ}$). Drift correction was performed after each text passage.

\section{Preprocessing}

In this Section, we describe the processing from the raw data recordings to the extraction of character-level and word-level reading time features. We share the following versions of the data: raw, character-level fixation information, character-level saccade information, and word-level eye tracking features.

\subsection{Gaze Event Detection}
A fixation is defined as a time window during which the eye is relatively still and focuses on the same point. For CopCo, the eye movement events are generated in real-time by the EyeLink eye tracker software during recording with a velocity- and acceleration-based saccade detection method.  A fixation event is defined by the algorithm as any period that is not a saccade or a blink. Hence, the raw data consist of (x,y) gaze location coordinates for individual fixations.

We use the DataViewer software by SR Research to extract fixation events for all areas of interest. Areas of interest are automatically defined as rectangular boxes that surround each individual character of a text on the screen, as shown in Figure \ref{fig:fix-saccs}. For later analysis, only fixations within the boundaries of each displayed character are extracted. Therefore, data points distinctly not associated with reading are excluded. An example of the resulting fixations and saccades is shown in Figure \ref{fig:fix-saccs}.

\begin{figure*}[t]
\centering
\includegraphics[width=0.99\textwidth]{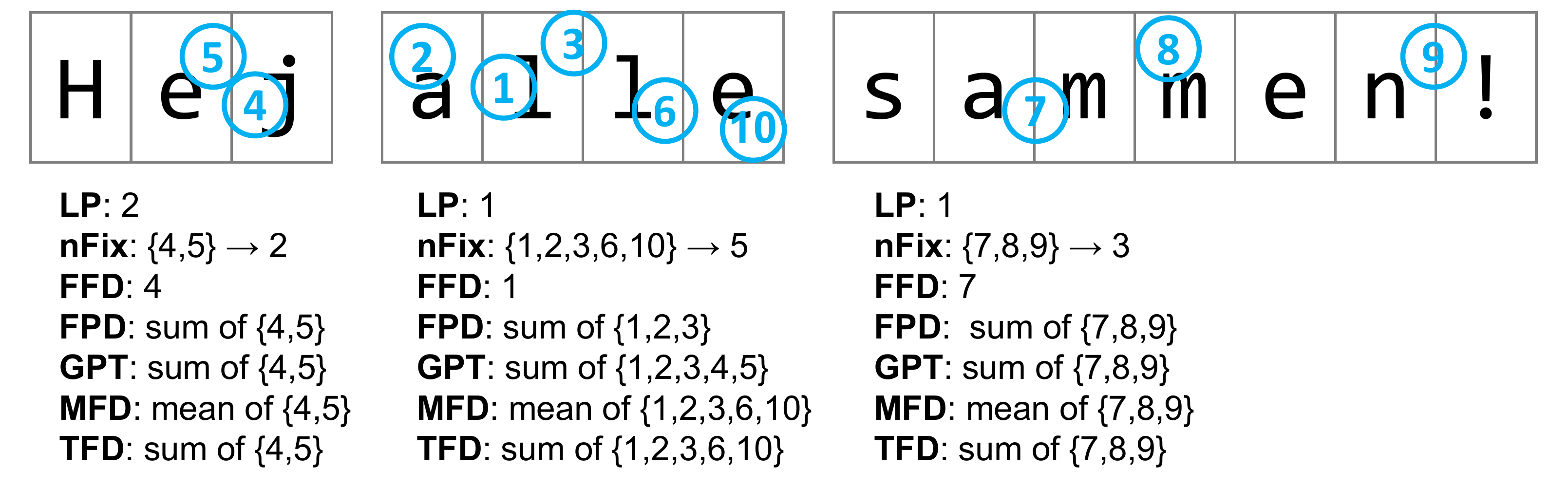}
\caption{Character-level to word-level feature mapping. The blue circles represent individual fixations and their order, and the gray boxes mark the character-level interest areas. Abbreviations: Landing position (LP), number of fixations (nFix), first fixation duration (FFD), first pass duration (FPD), go-past time (GPT), total fixation duration (TFD), mean fixtion duration (MFD).}\label{fig:feature-extraction}
\end{figure*}

\subsection{Feature Extraction} 
\label{sec:Feature_extraction}
In a second step, we use custom Python code to map and aggregate character-level features to word-level features. Figure \ref{fig:feature-extraction} depicts this process. To the best of our knowledge, we are the first to share such an automatic conversion script.\footnote{The code is available here: \url{https://github.com/norahollenstein/copco-processing}}

We extract the following eye tracking features:
\begin{enumerate}
    \item \textit{Number of fixations}, the total amount of fixations on the current word, including all passes.
    \item \textit{First fixation duration}, the duration (in milliseconds) of the first fixation on the prevailing word.
    \item \textit{Mean fixation duration}, the sum of all fixation durations (in milliseconds) on the current word divided by the number of fixations.
    \item \textit{Total fixation duration}, the sum of all fixation durations (in milliseconds) on the current word.
    \item \textit{First pass duration}, the summed duration (in milliseconds) of all fixations on the current word prior to progressing out of the current word (to the left or right). 
    \item \textit{Go-past time}, the sum duration (in milliseconds) of all fixations prior to progressing to the right of the current word, including regressions to previous words that originated from the current word.
    \item \textit{Landing position}, index of the first character fixation in the prevailing word (if the fixation falls into multiple interest areas, the character on the left is chosen).
    \item \textit{Mean saccade duration}, the mean duration (in milliseconds) of all saccades originating from the current word.
    \item \textit{Peak saccade velocity}: Maximum gaze velocity (in visual degrees per second) of all saccades originating from the current word.

\end{enumerate}
These features are defined to cover the reading process from early lexical access to later syntactic integration.
The selection of features is inspired by similar corpora in other languages \cite{hollenstein2018zuco,cop2017presenting} and extended to include character-level features, which will enable more fine-grained psycholinguistic analyses (i.e., differences in landing positions and character types between languages and scripts).

\section{Data Validation}
To ensure the quality of the recorded data, we present a series of analyzes that take a closer look at reading comprehension, the effects of word length and word frequency, the effects on landing position at character level, and a comparison of the extracted features.

\paragraph{Reading comprehension.}
Based on the scores of the reading comprehension questions of all participants (as presented in Table \ref{tab:participants}), the mean accuracy is 82\%, with a minimum of 62\% and a maximum of 100\%. Therefore, no participant data needs to be excluded due to low comprehension. On average, participants read four speeches in a one-hour session. The mean reading time per screen is 17.65 seconds ($\pm3.84$). All except one participants fall within two standard deviations of the mean. In this participant population, there is no significant correlation between the reading comprehension scores and the reading times per screen (Spearman's rank correlation coefficient: 0.1, $p>0.6$).

\begin{figure}[t]
\includegraphics[width=0.5\textwidth]{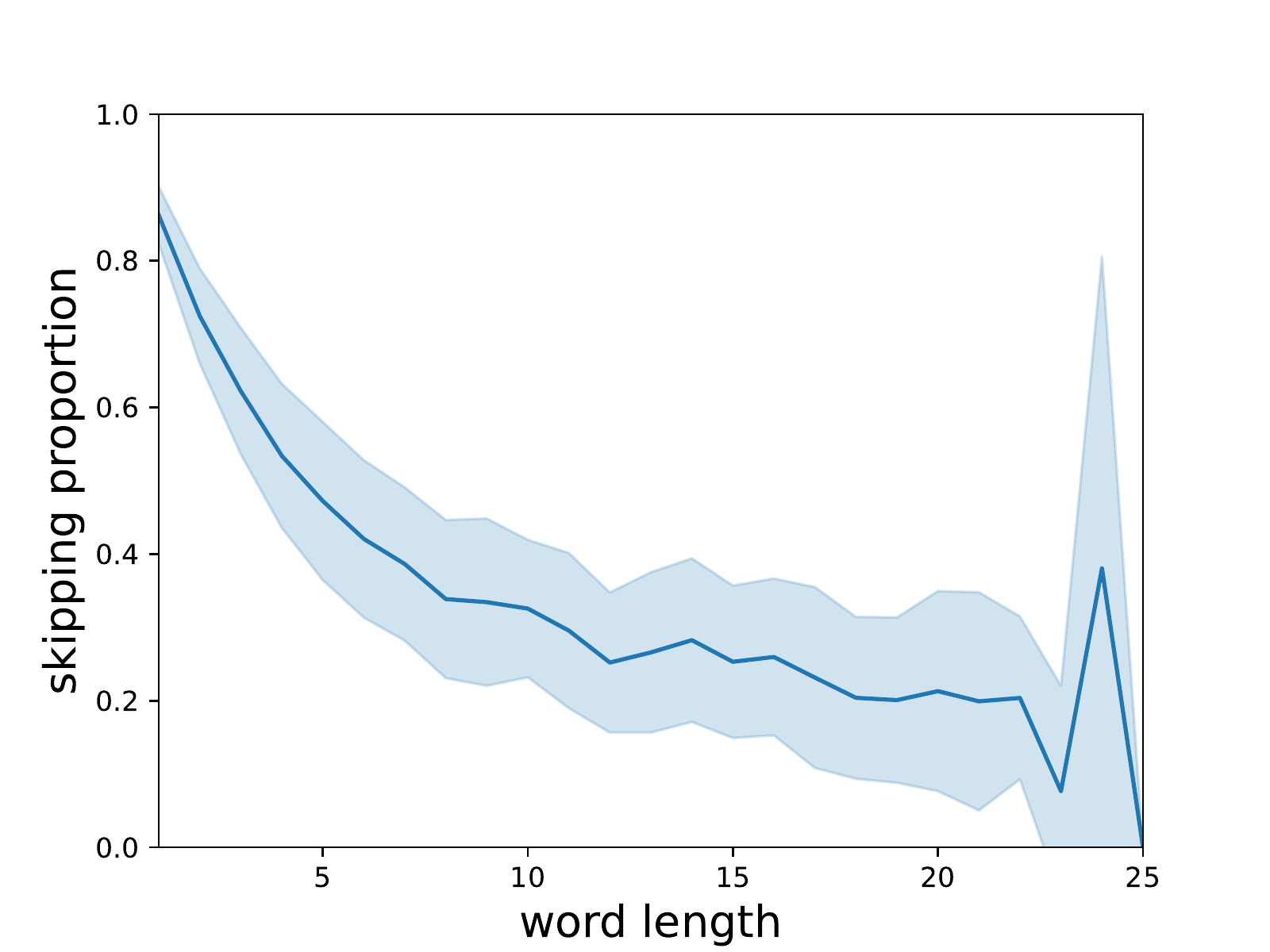}
\caption{Effect of word length on the skipping proportion across all participants (i.e., the proportion of readers that fixate a given word), with the standard deviation in the shaded area.}\label{fig:word-legnth}
\end{figure}

\begin{figure}[t]
\includegraphics[width=0.5\textwidth]{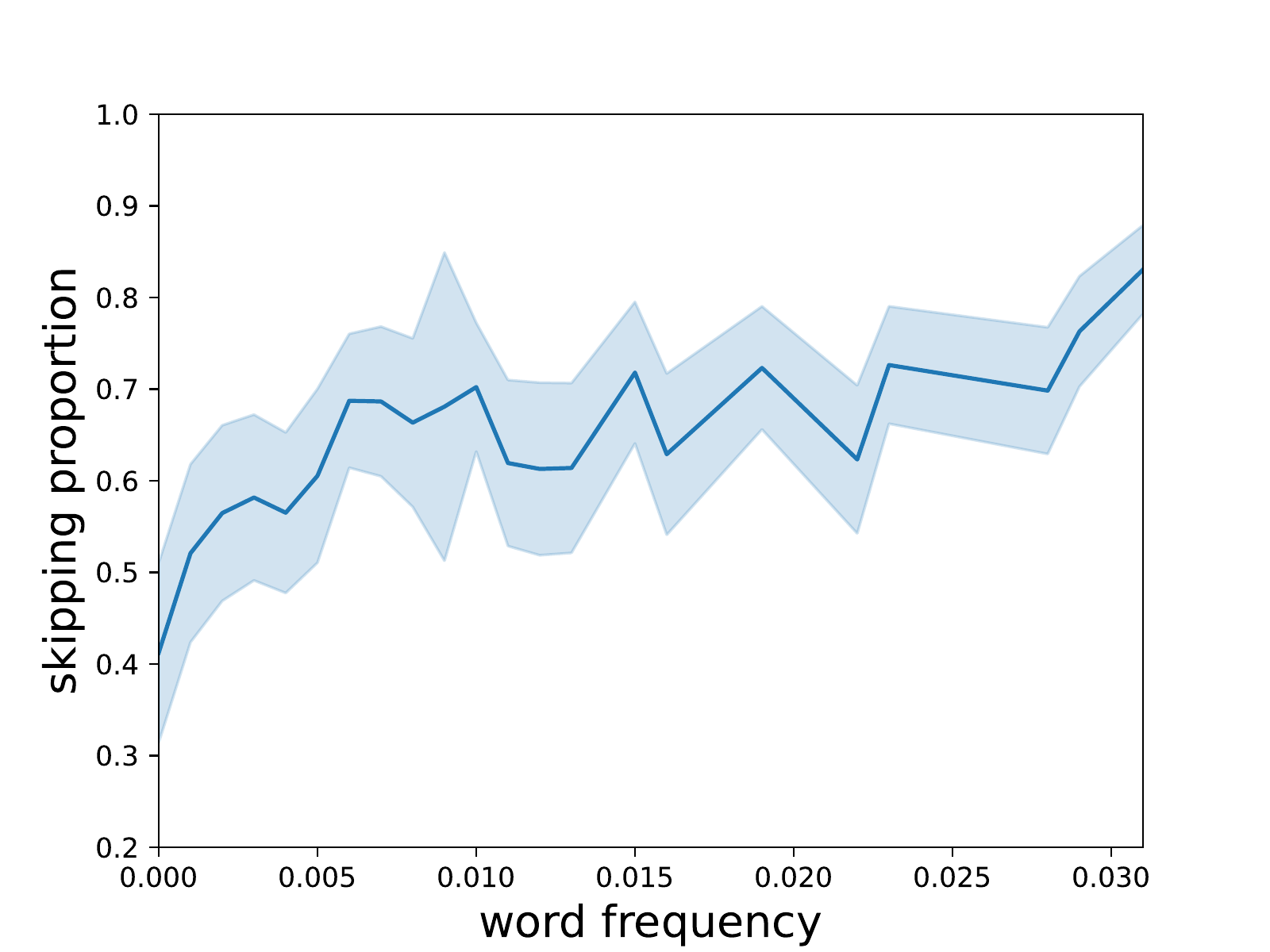}
\caption{Effect of word frequency on the skipping proportion across all participants (i.e., the proportion of readers that fixate a given word), with the standard deviation in the shaded area.}\label{fig:word-freq}
\end{figure}

\paragraph{Word length \& word frequency.}
Eye movements during reading are regulated by various lexical aspects such as word length and word frequency: Longer and less frequent words are more likely to be fixated.
Furthermore, these word characteristics affect fixation duration similarly across languages, but the size of the effect depends on
the language and the script \cite{laurinavichyute2019russian,bai2008reading}. As we can observe in the new CopCo corpus, this is also the case in Danish. In Figure \ref{fig:word-legnth}, we show the effect of word length found in the eye tracking data recorded from the Danish stimulus. Figure \ref{fig:word-freq} shows how more frequent words in the Danish language are skipped more often in the recorded eye tracking data. The general skipping rate for participants (i.e., the proportion of words that are not fixated) lies between 0.29 and 0.58.

\paragraph{Landing position.}
Next, we analyze the landing position within a word.
In early research, \newcite{liversedge1998foveal} suggested that orthographic information, such as the frequency of characters visible in the parafovea, may influence the landing position on the following word. In the CopCo data, we find that the number of times a character is fixated first within a word correlates only moderately with the character frequency in Danish (Spearman's rank correlation coefficient: 0.35, $p=0.069$). \newcite{rayner1992eye} concluded that low-level visual information (primarily word length) is the key determinant of the initial landing position on a word during reading. We find this effect in the CopCo data: The character landing position index highly correlates with the word length, meaning that for longer words the gaze tends to land on later characters (Spearman's rank correlation coefficient: 0.90, $p<0.0001$). Finally, the fixation duration on the landing character does not differ significantly between vowels and consonants.

\begin{figure}[t]
\centering
\includegraphics[width=0.5\textwidth]{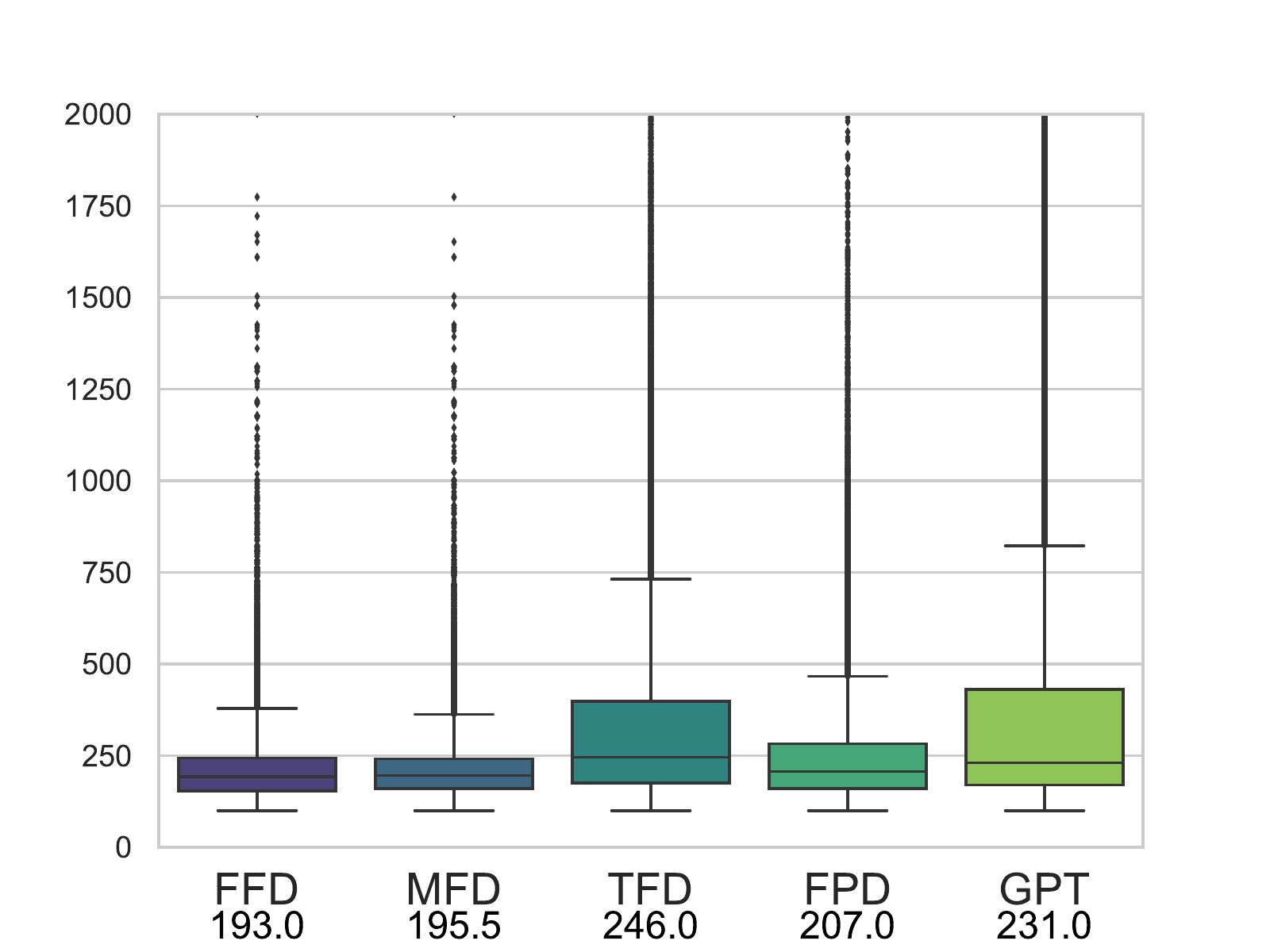}
\caption{Feature ranges for the first fixation duration (FFD), mean fixation duration (MFD), total fixation duration (TFD), first-pass duration (FPD), and go-past time (GPT) with the median feature value below each boxplot.}\label{fig:feature-ranges}
\end{figure}

\paragraph{Feature ranges.} 
Lastly, we compare the extracted word-level eye tracking features to existing corpora. Fixations shorter than 100 ms were excluded from the analysis, because these are unlikely to reflect fixations relevant for reading \cite{sereno2003measuring}. On average, a word is fixated 0.69 times (standard deviation: 1.05). Figure \ref{fig:feature-ranges} shows the mean and range of each of the five most commonly extracted fixation features: first fixation duration, mean fixation duration, total fixation duration, first-pass duration and go-past time. These values are in line with similar corpora in other languages, e.g., the GECO corpus \cite{cop2017presenting} and the ZuCo corpus \cite{hollenstein2018zuco}.

\section{Outlook}

CopCo can be used for applications in the fields of natural language processing as well as in psycholinguistics and reading research.
In NLP, eye movement features can be leveraged to improve models for syntactic and semantic language understanding tasks (e.g., part-of-speech tagging \cite{barrett2016weakly,klerke-plank-2019-glance} or named entity recognition \cite{hollenstein2019entity}). This new eye tracking dataset also allows us to analyze and interpret language models or task-specific NLP models. For example, we can investigate machine-learning based explainability mechanisms such as attention and saliency in Danish language models, as suggested by \newcite{hollenstein-beinborn-2021-relative} or \newcite{sood2020interpreting} for English.

In psycholinguistics, the CopCo data can be used to study human reading, including the analysis of reading patterns of Danish native speakers, investigating differences between individual readers or subgroups of readers (e.g., split by age or gender), and the prediction of eye movements from reading Danish texts to develop more accurate reading models. Moreover, it enables further exploration of linguistic phenomena in natural reading, e.g., processing of relative clauses or negation. Finally, this new Danish eye tracking corpus allows cross-linguistic analysis of eye movements while reading by comparing with available eye tracking corpora in other languages. 


CopCo is designed to be a growing corpus. Future releases will include (i) recordings from additional participant populations such as dyslexic readers and Danish language learners, and (ii) reading materials of other text genres, for instance, Wikipedia articles or social media posts.

\section*{Acknowledgements}
Maria Barrett is supported by a research grant (34437) from VILLUM FONDEN. 

\section{Bibliographical References}\label{reference}

\bibliographystyle{lrec2022-bib}
\bibliography{lrec2022-example,cog-sci-nlp-2021}

\end{document}